\documentclass[11pt]{article}
\usepackage[final]{acl}
\usepackage{times}
\usepackage{latexsym}
\usepackage[T1]{fontenc}
\usepackage[utf8]{inputenc}
\usepackage{microtype}
\usepackage{booktabs}
\usepackage{graphicx}
\usepackage{xcolor}
\usepackage{url}
\usepackage{amsmath}
\usepackage{enumitem}
\newcommand{\code}[1]{\texttt{#1}}

\usepackage{placeins}

\newcommand{\FirstSeatOne}{51.5 [48.6, 54.3]}

\newcommand{\NRowsFile}{3{,}172}

\newcommand{\CallsAll}{61{,}571}
\newcommand{\CallsLocal}{19{,}301}
\newcommand{\FirstSeatAll}{52.3 [50.5, 54.2]}
\newcommand{\NDecidedAll}{2{,}828}
\newcommand{\OccupiedShareLlamaFree}{75}
\newcommand{\OccupiedShareMinFree}{43}
\newcommand{\WinSearchGreedy}{91.0 [86.2, 94.2]}

\newcommand{\WinSearchRule}{95.0 [91.0, 97.3]}

\newcommand{\WinSearchLook}{89.0 [83.9, 92.6]}

\newcommand{\WinSearchRandom}{100.0 [98.1, 100.0]}

\newcommand{\WinAGreedyRule}{65.0}
\newcommand{\WinBGreedyRule}{32.0}

\newcommand{\WinARuleLook}{69.0}
\newcommand{\WinBRuleLook}{30.0}

\newcommand{\WinALookRandom}{97.0}
\newcommand{\WinBLookRandom}{3.0}

\newcommand{\WinLlamaTypedRandom}{10.0 [4.3, 21.4]}

\newcommand{\RecLlamaTypedRandom}{5--37--8}

\newcommand{\WinMinTypedRandom}{2.0 [0.4, 10.5]}

\newcommand{\RecMinTypedRandom}{1--47--2}

\newcommand{\RecLlamaTypedLlamaFree}{41--5--4}

\newcommand{\RecMinTypedMinFree}{38--10--2}

\newcommand{\RecGemRule}{0--50}

\newcommand{\WinNimRule}{40.0 [27.6, 53.8]}

\newcommand{\RecNimRule}{20--27--3}

\newcommand{\RecGemLook}{11--38--1}

\newcommand{\RecNimLook}{32--18}

\newcommand{\RecGemRandom}{40--9--1}

\newcommand{\RecNimRandom}{50--0}

\newcommand{\WinLocLlamaTypedRandom}{4.0 [1.1, 13.5]}

\newcommand{\RecLocLlamaTypedRandom}{2--30--18}

\newcommand{\WinLocQwenTypedRandom}{30.0 [19.1, 43.8]}

\newcommand{\RecLocQwenTypedRandom}{15--20--15}

\newcommand{\WinLocLlamaShufRandom}{18.0 [9.8, 30.8]}

\newcommand{\RecLocLlamaShufRandom}{9--20--21}

\newcommand{\WinLocQwenShufRandom}{22.0 [12.8, 35.2]}

\newcommand{\RecLocQwenShufRandom}{11--28--11}

\newcommand{\CallsGameLlamaTyped}{53}

\newcommand{\CallsGameLlamaFree}{30}

\newcommand{\NDecGem}{11{,}459}

\newcommand{\TimeoutsGem}{4}
\newcommand{\CallsGameGem}{76}
\newcommand{\CallsTotalGem}{11{,}371}
\newcommand{\ForcedTotalGem}{88}

\newcommand{\TokensGameGem}{146{,}661}

\newcommand{\NDecNim}{21{,}706}

\newcommand{\TimeoutsNim}{2}
\newcommand{\CallsGameNim}{136}
\newcommand{\CallsTotalNim}{20{,}342}
\newcommand{\ForcedTotalNim}{1{,}367}

\newcommand{\TokensGameNim}{299{,}405}

\newcommand{\NDecLocLlamaTyped}{8{,}285}

\newcommand{\NDecLocQwenTyped}{9{,}565}

\newcommand{\NDecLocLlamaFree}{720}

\newcommand{\BadLocLlamaFree}{720}

\newcommand{\NDecLocQwenFree}{731}

\newcommand{\BadLocQwenFree}{720}

\newcommand{\NBatchFive}{800}
\newcommand{\MsDecSearch}{60.7}
\newcommand{\BadSearch}{0}

\newcommand{\BadTwoLlamaFree}{2{,}966/3{,}026}

\newcommand{\BadTwoMinFree}{2{,}983/3{,}032}

\newcommand{\MedMsGem}{367}

\newcommand{\MeanPGem}{0.22}

\newcommand{\PWinGem}{0.18}
\newcommand{\PLossGem}{0.24}

\newcommand{\MedMsNim}{348}

\newcommand{\MeanPNim}{0.16}

\newcommand{\PWinNim}{0.15}
\newcommand{\PLossNim}{0.20}
\newcommand{\NBatchSix}{272}
\newcommand{\SecDecLocLlamaTyped}{1.6}
\newcommand{\SecDecLocQwenTyped}{1.4}
\newcommand{\SecDecLocLlamaFree}{1.4}
\newcommand{\SecDecLocQwenFree}{2.3}

\newcommand{\SecDecBlockLocLlamaTyped}{1.4}
\newcommand{\SecDecBlockLocQwenTyped}{1.3}
\newcommand{\SecDecBlockRange}{1.3--1.4}

\newcommand{\ReplaySixGames}{272}
\newcommand{\ReplaySixMismatches}{0}

\newcommand{\ActLlamaFreeEndturn}{98}

\newcommand{\ActLlamaTypedEndturn}{52}
\newcommand{\ActLlamaTypedMove}{28}

\newcommand{\MedUnspentLlamaTyped}{20}
\newcommand{\DecGameLlamaTyped}{53}
\newcommand{\ActRandomVsTypedEndturn}{26}
\newcommand{\ActRandomVsTypedMove}{57}

\newcommand{\MedUnspentRandomVsTyped}{6}
\newcommand{\DecGameRandomVsTyped}{114}

\newcommand{\ActNimEndturn}{17}

\newcommand{\MedUnspentNim}{1}
\newcommand{\DecGameNim}{145}

\newcommand{\ActMinTypedEndturn}{71}
\newcommand{\ActMinTypedMove}{11}

\newcommand{\MedUnspentMinTyped}{32}
\newcommand{\DecGameMinTyped}{38}
\newcommand{\ReplayGames}{800}
\newcommand{\ReplayMismatches}{0}

\newcommand{\PosDecisions}{108{,}092}

\newcommand{\PosFirstDLocLlamaTyped}{72.5}

\newcommand{\PosEndLocLlamaTyped}{32.5}
\newcommand{\PosUnifLocLlamaTyped}{29.5}

\newcommand{\PosFirstDLocLlamaTypedVR}{66.2}

\newcommand{\PosUnifLocLlamaTypedVR}{27.1}

\newcommand{\PosEndLocQwenTyped}{28.7}

\newcommand{\PosFirstDLocQwenTypedVR}{42.5}

\newcommand{\PosUnifLocQwenTypedVR}{24.6}

\newcommand{\PosFirstDLocLlamaShuf}{62.4}
\newcommand{\PosLastDLocLlamaShuf}{15.5}
\newcommand{\PosEndLocLlamaShuf}{39.5}
\newcommand{\PosUnifLocLlamaShuf}{22.1}

\newcommand{\PosFirstDLocQwenShuf}{21.4}
\newcommand{\PosLastDLocQwenShuf}{10.4}
\newcommand{\PosEndLocQwenShuf}{46.7}
\newcommand{\PosUnifLocQwenShuf}{16.7}

\newcommand{\AptLlamaTyped}{0.9}
\newcommand{\AptEndLlamaTyped}{52}

\newcommand{\AptRandomHosted}{2.8}

\newcommand{\AptMinTyped}{0.4}
\newcommand{\AptEndMinTyped}{71}

\newcommand{\AptRandomLocal}{2.8}

\newcommand{\AptLocQwenTypedVR}{2.7}

\newcommand{\AptLocQwenShuf}{1.1}

\title{Legal Is Not Good: A Pre-Registered Benchmark of Low-Cost Language\\ and Decision Models on a New Strategy Game}

\author{Federico Lucca \\
  Independent researcher, Trento, Italy \\
  \texttt{federico.lucca@gmail.com}}

\begin{document}
\maketitle
\begin{abstract}
\emph{Measure Before You Train} (MBYT) argued that an agent's decisions should be measured against ground truth, under a protocol fixed in advance, before anyone trains or replaces the model behind them. We carry the method to a second, very different domain---a turn-based strategy game---and ask a deliberately modest question: how well do \emph{low-cost} agents play under a pre-registered protocol? hexfolk is a minimal two-player 4X game written for this purpose, so no model has seen its rules or positions. A deterministic engine, one agent interface, a referee that never corrects an illegal or late move, and per-decision logs make every game replayable; the questions and their expected answers were committed before the games, on held-out seeds with mirrored seats and 95\% Wilson intervals. Five free reference players rank exactly as pre-registered, topped by a 60\,ms-per-move search that beats every other one in at least 89\% of games. Two small open language models (8B and 3B) lose every game to a few-line \code{rule} player. Answering in free text they are invalid on 98\% of their decisions; constrained to legal moves they never play an invalid one, yet they lose to the \emph{uniformly random} player (the 8B model wins \WinLlamaTypedRandom\% of games), refuting our expectation that they would beat it; the result survives a control that shuffles the order of the options. Two native decision models also never play an invalid move and both beat \code{random}; one loses all 50 games to \code{rule}; the other is not distinguishable from \code{rule} in 50 games (\WinNimRule\% wins) and beats a one-step search. The paid agents make tens to over a hundred model calls per game; against the same opponents the free search, which makes none, outplays every paid model by the full range of the scale. Two 7--8B models served locally behave the same way: typed, the Llama model loses to \code{random} too (\WinLocLlamaTypedRandom\% wins) and the Qwen model is not distinguishable from it. No frontier model is included.
\end{abstract}

\section{Introduction}

Agents built on language models are increasingly asked to \emph{decide} rather than to write: which of a few dozen legal moves to play, which label to assign, whether to act. The companion paper~\citep{lucca2026mbyt} (MBYT) studied one such agent, a personal inbox router, and found that once ground truth existed a classical baseline beat every zero-shot model. Its method---freeze the inputs, fix the questions and the expected answers before looking, keep a held-out set, report intervals and refuse to rank what the intervals do not separate---is not specific to email, but a single domain cannot show that.

This paper is the second domain. A game makes the ground truth free and exact (the winner is not a matter of annotation), makes strong non-learned baselines cheap to write, and can be written from scratch: hexfolk was built in November 2026, so its rules, names and positions cannot be in any training set, which removes the data-contamination concern of benchmarks built on public games~\citep{sainz2023contamination}. Its mechanics borrow from a genre (Section~\ref{sec:limits}).

We ask the question at the centre of MBYT---do native \emph{decision models}, which return a probability distribution over caller-defined options instead of text~\citep{jev,nimble}, decide as well as text-generating language models and as well as simple hand-written players?---and we ask it of \emph{cheap} agents on purpose. The study works with a limited budget of paid API calls, which rules out frontier models and larger paid seed sets; what it allows is a careful answer for the agents a small team would deploy per move, set against players that make no model call at all.

\paragraph{Contributions.} (1) hexfolk and its benchmark harness: a deterministic engine, one agent interface for built-in players, language models, decision models and MCP clients, a referee that counts every invalid or late move, and replayable logs. (2) A pre-registered protocol whose amendments are all recorded before the games they govern. (3) Results for five free reference players (including a search ceiling), four small-language-model agents, two decision models and four locally served agents, with quality reported against cost, and a refuted expectation reported as such: an agent restricted to legal moves can still play worse than chance.

\section{The Game and the Harness}\label{sec:game}

\paragraph{Rules.} hexfolk (rules v1.0.0, frozen) is played on a seeded 11$\times$11 island of hexes with five terrains. Each player starts with a capital, one warrior and 5 stars; neutral villages wait to be captured. Cities produce stars, the only currency, which buy units (four types), six techs, harvests of adjacent resources (+1 population), roads and walls. At the start of each round two seeded dice pay a star for every harvested tile carrying that number. Combat is deterministic. A player wins by capturing the enemy capital or by the higher score after round 30; equal scores are a draw. There is no fog of war.

\paragraph{Engine and referee.} The rules are a pure TypeScript engine (\code{newGame}, \code{observe}, \code{legalActions}, \code{step}) with no randomness outside the seed, so a game is determined by its seed and action list. The browser game, the built-in AI, the headless benchmark and an MCP server~\citep{anthropic2024mcp} all play through it. In the benchmark an agent makes one action per decision, given the observation and the legal actions; the turn passes when it plays \code{end\_turn}. The referee never corrects a move: a malformed or illegal action, an API failure after three retries or no reply within 60\,s is logged, counted, and replaced by \code{end\_turn}; more than 20 actions in a turn ends it. Each game yields a row of \code{results.csv} and a JSONL log of every decision with its latency, raw reply and provider usage.

\paragraph{Agents.} Four reference players need no model: \code{random} picks uniformly among the legal actions; \code{rule} is a priority list of a few lines (capture, winning attack, keep units at cities, harvest, research, move towards the nearest city it does not own, end turn); \code{lookahead} tries every legal action on a copy of the state and keeps the best under a static evaluation that includes the opponent's next-turn damage; \code{greedy-normal} is the game's own tuned AI. The model agents read the same rules and state text (an observation summary, a terrain map, the unharvested resources). A \emph{free} agent writes one JSON action. A \emph{typed} agent sees the numbered legal actions and answers \code{\{"choice":"N"\}} under a JSON schema whose only allowed values are the legal indices. A \emph{decision} agent sends the numbered legal actions as the options of one \code{choice} question through the System One interface~\citep{jev} and plays the top-ranked option; it cannot produce a malformed or illegal action, except a position with more than 255 legal actions (\code{tooManyOptions}). Positions with a single legal action are played without a model call. Models: \code{llama8b} = Llama~3.1 8B Instruct and \code{ministral3b} = Ministral 3B (OpenRouter, temperature 0); \code{gemini-decision} = \code{google/gemini-decision-1} (OpenRouter; released 11 November 2026, \citealp{choi2026gemini}); \code{nimble-decision} = Bespoke Nimble v4-max~\citep{nimble}.

\section{Pre-Registered Protocol}\label{sec:protocol}

The protocol (\code{docs/PROTOCOL.md}) writes the questions and expected answers before the numbers exist~\citep{nosek2018prereg}. \emph{Tuning seeds} (1--1{,}000) are for development and pilots; \emph{test seeds} 100001--100100 are held out and played once, every pairing from both seats; a batch that must be rerun is rerun whole and both runs are reported. Every rate has a 95\% Wilson interval~\citep{wilson1927}; A is better than B only if A's win-rate interval lies entirely above B's, otherwise they are \emph{not distinguishable}. Batch~1 plays the six reference pairings on all 100 test seeds (1{,}200 games); the paid batches use seeds 100001--100025 (50 games per pairing): batch~2, the four language-model agents against \code{random} and \code{rule}; batch~3, typed against free for each model; batch~4, each decision agent against \code{random}, \code{rule} and \code{lookahead}. Version~1 was committed before any result; batch~1 ran under it before amendment~v2 added the decision agents and batch~4; v2.1 records, before any language-model test game, a forced-move fix, a stop at the first out-of-credit error, a spending limit and one prompt rule found on tuning seeds (a city can train only on an empty tile); v3 (19 Nov 2026), written after batches 1--4, adds a free search player and batch~5 (Section~\ref{sec:search}); v4 and v4.1 (20 Nov 2026), written before any local game, add locally served models and batch~6 (Q10--Q16), v4.2 records that the owner cut batch~6 at 6 of its 25 seeds, and v4.3, committed before the extra games, continued only the typed local agents against \code{random} and \code{rule} to all 25 seeds (Section~\ref{sec:local}); v5 (20 Nov 2026, committed before any test game of it) adds an option-order control, batch~7, and Q17--Q19 (Section~\ref{sec:order}). Appendix~\ref{app:history} gives the full history with commits. The paid batches made \CallsAll{} model calls in all; the local batch~6 made \CallsLocal{}, none of them paid.

\begin{table*}[t]
\centering\small
\caption{Pre-registered questions, expectations and status. Batches 1--5 and 7 are final (batch~4: run~2). Batch~6 (Q10--Q16) is reduced: the typed local agents have 50 games against \code{random} and \code{rule}; the free local agents and the games against \code{search} have only 12 (indicative): Q14 and those parts of Q10 and Q13 are in Appendix~\ref{app:batch6}.}
\label{tab:questions}
\footnotesize\setlength{\tabcolsep}{3pt}
\begin{tabular}{p{0.05\textwidth}p{0.22\textwidth}p{0.30\textwidth}p{0.37\textwidth}}
\toprule
\# & Question & Expectation (written in advance) & Status \\
\midrule
Q1 & Typed vs.\ free agent, same model? & Typed better than free, both models. & \textbf{Confirmed} for both (\RecLlamaTypedLlamaFree, \RecMinTypedMinFree). \\
Q2 & Does a language model beat \code{rule}? & No: \code{rule} better or not distinguishable, all 4 agents. & \textbf{Confirmed}: \code{rule} better than all 4 (50--0 each). \\
Q3 & Does a language model beat \code{random}? & Typed: better. Free: not distinguishable. & \textbf{Refuted}, both parts: \code{random} is better than all 4 agents. \\
Q4 & Invalid moves per agent? & References 0; typed $<$1\%; free 10--50\%. & References and typed: \textbf{confirmed} (0). Free: \textbf{not confirmed}, 98\%. \\
Q5 & 8B vs.\ 3B model? & 8B at least as high, not distinguishable. & \textbf{Confirmed} (both near zero). \\
Q6 & Ranking of reference players? & \code{greedy-normal} $>$ \code{rule} $>$ \code{lookahead} $>$ \code{random}, each step distinguishable. & \textbf{Confirmed} (Table~\ref{tab:pairs}). \\
Q7 & Does seat order matter? & First-seat win rate 45--60\%. & \textbf{Confirmed}: \FirstSeatOne\% (batch~1); \FirstSeatAll\% (batches 1--5). \\
Q8 & Decision model vs.\ typed agents and references? & Not distinguishable from typed agents; better than \code{random}; \code{rule} and \code{lookahead} better or not distinguishable. & References: \textbf{confirmed} for gemini; nimble \textbf{not confirmed} vs.\ \code{lookahead} (it is better). Typed part: \textbf{not confirmed} (both decision agents better). \\
Q9 & Invalid moves of a decision agent? & 0 apart from \code{tooManyOptions}, those $<$1\%. & \textbf{Confirmed} for both (0, and no \code{tooManyOptions}). \\
Q10s & Is \code{search} a ceiling above the other references? (v3) & \code{search} better than each other reference (win-rate CI above 50\%). & \textbf{Confirmed}: lowest interval \WinSearchLook\% (vs.\ \code{lookahead}). \\
Q10 & Local vs.\ hosted \code{llama8b} (v4)? & Win rates vs.\ \code{random} and \code{rule}, and invalid rates, not distinguishable. & Partly: win rates \textbf{confirmed} (typed vs.\ \code{random} \WinLocLlamaTypedRandom\% local vs.\ \WinLlamaTypedRandom\% hosted); typed invalid rate confirmed (0 vs.\ 0); free agents: Appendix~\ref{app:batch6}. Direct match not played. \\
Q11 & Do local agents replay identically? & Yes; exploratory seed replays. & From the log: yes (all \ReplaySixGames{} games). From the seed: not tested on test seeds, not supported by the pilot. \\
Q12 & Local \code{qwen7b} vs.\ \code{llama8b}, typed? & Not distinguishable vs.\ \code{random} and \code{rule}. & Partly: vs.\ \code{rule} confirmed (0\% each); vs.\ \code{random} \textbf{refuted}: Qwen is better (\WinLocQwenTypedRandom\% vs.\ \WinLocLlamaTypedRandom\%, disjoint). \\
Q13 & Local typed agents? & Better than \code{random}; \code{rule} and \code{search} better. & Partly: \code{rule} better (0 of 100); ``better than \code{random}'' \textbf{refuted} for Llama (\code{random} better) and not confirmed for Qwen (n.d.); \code{search}: Appendix~\ref{app:batch6}. \\
Q14 & Local free agents (12 games). & & Appendix~\ref{app:batch6}. \\
Q15 & Local decision model. & Dropped before testing (no local server for it). & -- \\
Q17 & Does option order drive the \code{end\_turn} choices? (v5) & With shuffled order the \code{end\_turn} share falls by $\geq$10 points, both local models. & \textbf{Refuted}, opposite direction: it rises (\PosEndLocLlamaTyped\%$\to$\PosEndLocLlamaShuf\%, \PosEndLocQwenTyped\%$\to$\PosEndLocQwenShuf\%). \\
Q18 & Does option order make the typed agents lose? & Win rate vs.\ \code{random} with shuffled order n.d.\ from batch~6. & \textbf{Confirmed}: \WinLocLlamaShufRandom\% vs.\ \WinLocLlamaTypedRandom\%; \WinLocQwenShufRandom\% vs.\ \WinLocQwenTypedRandom\%. \\
Q19 & Position bias (exploratory). & First and last index vs.\ uniform, both orders. & Primacy for Llama in both orders; weaker for Qwen; no recency. \\
Q16 & Cost of local agents? & No paid call; report latency per decision. & Confirmed: \SecDecBlockRange\,s per typed decision with one model on the GPU; 5--15\,s when models swap. \\
\bottomrule
\end{tabular}
\end{table*}

\section{Results}\label{sec:results}

All numbers are produced by \code{data/analyze.py} from a copy of \code{results.csv}; tables and in-text figures are generated, not typed. The script reads a batch only when every pre-registered game in it is present; in the copies used here (\NRowsFile{} rows) batches 1--5 are complete. Batch~6 is read as protocols v4.2--v4.3 define it: the typed local agents against \code{random} and \code{rule} on all 25 seeds, every other pairing on seeds 100001--100006; its 12-game pairings are marked indicative wherever they appear. Batch~7 (the order control) is complete.

\begin{table*}[t]
\centering\small
\caption{Head to head on the test seeds, both seats. In paid batches the paid agent is A (in batch~3, the typed agent). A/B wins in \% with 95\% Wilson intervals; D = draws; verdict by non-overlapping intervals. Batch~5 (\code{search}, the ceiling) is listed first. Batch~6 lists its 50-game pairings; its 12-game pairings are in Appendix~\ref{app:batch6}. Batch~7 is the option-order control (Section~\ref{sec:order}).}
\label{tab:pairs}
\scriptsize\setlength{\tabcolsep}{4pt}\renewcommand{\arraystretch}{0.92}
\begin{tabular}{llrcccl}
\toprule
A & B & $n$ & A wins \% [95\% CI] & B wins \% [95\% CI] & D & Verdict \\
\midrule
\multicolumn{7}{l}{\emph{Batch 5 (free search ceiling, protocol v3, seeds 100001--100100)}} \\
\code{search} & \code{greedy-normal} & 200 & 91.0 [86.2, 94.2] & 8.5 [5.4, 13.2] & 1 & \code{search} better \\
\code{search} & \code{rule} & 200 & 95.0 [91.0, 97.3] & 4.5 [2.4, 8.3] & 1 & \code{search} better \\
\code{search} & \code{lookahead} & 200 & 89.0 [83.9, 92.6] & 8.0 [5.0, 12.6] & 6 & \code{search} better \\
\code{search} & \code{random} & 200 & 100.0 [98.1, 100.0] & 0.0 [0.0, 1.9] & 0 & \code{search} better \\
\midrule
\multicolumn{7}{l}{\emph{Batch 1 (free references, seeds 100001--100100)}} \\
\code{greedy-normal} & \code{rule} & 200 & 65.0 [58.2, 71.3] & 32.0 [25.9, 38.8] & 6 & \code{greedy-normal} better \\
\code{greedy-normal} & \code{lookahead} & 200 & 71.5 [64.9, 77.3] & 27.0 [21.3, 33.5] & 3 & \code{greedy-normal} better \\
\code{greedy-normal} & \code{random} & 200 & 98.0 [95.0, 99.2] & 2.0 [0.8, 5.0] & 0 & \code{greedy-normal} better \\
\code{rule} & \code{lookahead} & 200 & 69.0 [62.3, 75.0] & 30.0 [24.1, 36.7] & 2 & \code{rule} better \\
\code{rule} & \code{random} & 200 & 98.0 [95.0, 99.2] & 2.0 [0.8, 5.0] & 0 & \code{rule} better \\
\code{lookahead} & \code{random} & 200 & 97.0 [93.6, 98.6] & 3.0 [1.4, 6.4] & 0 & \code{lookahead} better \\
\midrule
\multicolumn{7}{l}{\emph{Batch 2 (language models vs.\ references, seeds 100001--100025)}} \\
\code{llama8b-typed} & \code{rule} & 50 & 0.0 [0.0, 7.1] & 100.0 [92.9, 100.0] & 0 & \code{rule} better \\
\code{ministral3b-typed} & \code{rule} & 50 & 0.0 [0.0, 7.1] & 100.0 [92.9, 100.0] & 0 & \code{rule} better \\
\code{llama8b-free} & \code{rule} & 50 & 0.0 [0.0, 7.1] & 100.0 [92.9, 100.0] & 0 & \code{rule} better \\
\code{ministral3b-free} & \code{rule} & 50 & 0.0 [0.0, 7.1] & 100.0 [92.9, 100.0] & 0 & \code{rule} better \\
\code{llama8b-typed} & \code{random} & 50 & 10.0 [4.3, 21.4] & 74.0 [60.4, 84.1] & 8 & \code{random} better \\
\code{ministral3b-typed} & \code{random} & 50 & 2.0 [0.4, 10.5] & 94.0 [83.8, 97.9] & 2 & \code{random} better \\
\code{llama8b-free} & \code{random} & 50 & 0.0 [0.0, 7.1] & 100.0 [92.9, 100.0] & 0 & \code{random} better \\
\code{ministral3b-free} & \code{random} & 50 & 0.0 [0.0, 7.1] & 100.0 [92.9, 100.0] & 0 & \code{random} better \\
\midrule
\multicolumn{7}{l}{\emph{Batch 3 (typed vs.\ free, same model, seeds 100001--100025)}} \\
\code{llama8b-typed} & \code{llama8b-free} & 50 & 82.0 [69.2, 90.2] & 10.0 [4.3, 21.4] & 4 & \code{llama8b-typed} better \\
\code{ministral3b-typed} & \code{ministral3b-free} & 50 & 76.0 [62.6, 85.7] & 20.0 [11.2, 33.0] & 2 & \code{ministral3b-typed} better \\
\midrule
\multicolumn{7}{l}{\emph{Batch 4 run 2 (decision models, seeds 100001--100025)}} \\
\code{gemini-decision} & \code{rule} & 50 & 0.0 [0.0, 7.1] & 100.0 [92.9, 100.0] & 0 & \code{rule} better \\
\code{nimble-decision} & \code{rule} & 50 & 40.0 [27.6, 53.8] & 54.0 [40.4, 67.0] & 3 & n.d. \\
\code{gemini-decision} & \code{lookahead} & 50 & 22.0 [12.8, 35.2] & 76.0 [62.6, 85.7] & 1 & \code{lookahead} better \\
\code{nimble-decision} & \code{lookahead} & 50 & 64.0 [50.1, 75.9] & 36.0 [24.1, 49.9] & 0 & \code{nimble-decision} better \\
\code{gemini-decision} & \code{random} & 50 & 80.0 [67.0, 88.8] & 18.0 [9.8, 30.8] & 1 & \code{gemini-decision} better \\
\code{nimble-decision} & \code{random} & 50 & 100.0 [92.9, 100.0] & 0.0 [0.0, 7.1] & 0 & \code{nimble-decision} better \\
\midrule
\multicolumn{7}{l}{\emph{Batch 6 (local typed agents, v4--v4.3, seeds 100001--100025)}} \\
\code{localllama8b-typed} & \code{rule} & 50 & 0.0 [0.0, 7.1] & 100.0 [92.9, 100.0] & 0 & \code{rule} better \\
\code{localqwen7b-typed} & \code{rule} & 50 & 0.0 [0.0, 7.1] & 100.0 [92.9, 100.0] & 0 & \code{rule} better \\
\code{localllama8b-typed} & \code{random} & 50 & 4.0 [1.1, 13.5] & 60.0 [46.2, 72.4] & 18 & \code{random} better \\
\code{localqwen7b-typed} & \code{random} & 50 & 30.0 [19.1, 43.8] & 40.0 [27.6, 53.8] & 15 & n.d. \\
\midrule
\multicolumn{7}{l}{\emph{Batch 7 (option-order control: local typed agents, shuffled order, v5, seeds 100001--100025)}} \\
\code{localllama8b-typed-shuffled} & \code{random} & 50 & 18.0 [9.8, 30.8] & 40.0 [27.6, 53.8] & 21 & n.d. \\
\code{localqwen7b-typed-shuffled} & \code{random} & 50 & 22.0 [12.8, 35.2] & 56.0 [42.3, 68.8] & 11 & \code{random} better \\
\bottomrule
\end{tabular}

\end{table*}

\paragraph{Reference players (batch 1).} Table~\ref{tab:pairs} answers Q6 as pre-registered: \code{greedy-normal} beats \code{rule} (\WinAGreedyRule\% vs.\ \WinBGreedyRule\%), \code{rule} beats \code{lookahead} (\WinARuleLook\% vs.\ \WinBRuleLook\%) and \code{lookahead} beats \code{random} (\WinALookRandom\% vs.\ \WinBLookRandom\%), every pairing with disjoint intervals, reproducing the tuning-seed table. A one-step search with a hand-tuned evaluation loses clearly to a few-line priority list: in hexfolk a coherent plan (expand, capture, keep a garrison) beats greedy evaluation, so \code{rule} is a demanding bar. No reference player makes an invalid move (Q4) and the first seat wins \FirstSeatOne\% of decided games in batch~1 and \FirstSeatAll\% of all \NDecidedAll{} decided games in batches 1--5 (Q7). Per-agent statistics and a win-rate figure are in Appendix~\ref{app:more}.

\paragraph{Small language models (batches 2--3).} Answering in free text, both models are almost always invalid: \BadTwoLlamaFree{} (\code{llama8b-free}) and \BadTwoMinFree{} (\code{ministral3b-free}) of their batch-2 decisions were malformed or illegal, each ending the turn, and they won no game. Even with the occupied-city rule now stated in the prompt, \OccupiedShareLlamaFree\% and \OccupiedShareMinFree\% of their invalid moves are attempts to train in an occupied city. Constrained to legal moves, the typed agents made no invalid move at all, yet they lost every game to \code{rule} and most of their games to \code{random}: \code{llama8b-typed} wins \WinLlamaTypedRandom\% (\RecLlamaTypedRandom{} win--loss--draw) and \code{ministral3b-typed} \WinMinTypedRandom\% (\RecMinTypedRandom). Typed beats free for both models (\RecLlamaTypedLlamaFree{} and \RecMinTypedMinFree), but only because the free agents hardly act.

\paragraph{Q3 is refuted.} We pre-registered that the typed agents would be better than \code{random} and the free agents not distinguishable from it. Instead \code{random} is better than all four, with disjoint intervals: a small language model that is only allowed legal moves plays this game worse than choosing among those moves at random. The other expectations hold or fail as follows: Q1 and Q2 are confirmed; Q4 is confirmed for the typed agents and not for the free ones, whose 98\% invalid rate is far above the expected 10--50\% even after the v2.1 prompt rule; Q5 is confirmed only in the weak sense that both models are near zero everywhere. The expectations are not revised; the refutation is the result.

\paragraph{Why (exploratory): short turns.} Replaying the logs through the engine (\ReplayGames{} games, \ReplayMismatches{} mismatches; Appendix~\ref{app:actions}) shows that the typed agents act little per turn. \code{end\_turn} is the only way to end a turn, so it is chosen about once per turn whatever else happens, and its share of an agent's actions is about $1/(a+1)$, where $a$ is the number of other actions per turn: a high share means short turns, not a liking for the option. The hosted typed agents' \code{end\_turn} shares (\AptEndLlamaTyped\% and \AptEndMinTyped\%) correspond to \AptLlamaTyped{} and \AptMinTyped{} actions per turn, against \AptRandomHosted{} for \code{random} in the same games (Table~\ref{tab:apt}): they pass with units at home and stars unspent. Short turns are not the whole story, though: \code{localqwen7b-typed} takes \AptLocQwenTypedVR{} actions per turn against \code{random}, about as many as \code{random} itself (\AptRandomLocal), and still does not beat it. This is a description, not a finding of the protocol.

\paragraph{Order control (batch 7).}\label{sec:order} In batches 2--6 the engine appends \code{end\_turn} after every other legal action and the prompt numbers the actions $0..n{-}1$ in that order, so \code{end\_turn} is always the last option (all \PosDecisions{} rebuilt decisions), and the typed agents choose strongly by position (Appendix~\ref{app:actions}). To test whether this order is why they lose, protocol v5 (expectations written and committed before any test game) ran batch~7: the two local typed agents against \code{random} on seeds 100001--100025, both seats, with the list shown in a seeded random order at every decision (the permutation is logged, and every logged choice maps back to the applied action). The order is not the reason. Win rates against \code{random} with shuffled order are not distinguishable from batch~6 (Q18 confirmed): \code{localllama8b-typed} \WinLocLlamaShufRandom\% (\RecLocLlamaShufRandom) against \WinLocLlamaTypedRandom\%, \code{localqwen7b-typed} \WinLocQwenShufRandom\% (\RecLocQwenShufRandom) against \WinLocQwenTypedRandom\%. And Q17 is refuted in the opposite direction: the \code{end\_turn} share does not fall when \code{end\_turn} stops being last, it rises (\PosEndLocLlamaTyped\% to \PosEndLocLlamaShuf\% for Llama, \PosEndLocQwenTyped\% to \PosEndLocQwenShuf\% for Qwen), because shuffled Qwen plays fewer actions per turn (\AptLocQwenShuf{} against \AptLocQwenTypedVR). A position bias does exist (Q19): Llama picks the first shown option in \PosFirstDLocLlamaTypedVR\% and \PosFirstDLocLlamaShuf\% of its decisions (engine order vs.\ \code{random}, shuffled) against \PosUnifLocLlamaTypedVR\% and \PosUnifLocLlamaShuf\% for a uniform choice; Qwen \PosFirstDLocQwenTypedVR\% and \PosFirstDLocQwenShuf\% against \PosUnifLocQwenTypedVR\% and \PosUnifLocQwenShuf\%; and the last shown option is below uniform when shuffled (\PosLastDLocLlamaShuf\% and \PosLastDLocQwenShuf\%), so there is no recency bias. Legal-only small models lose to \code{random} whatever order the options come in.

\paragraph{Decision models (batch 4).} A first run was invalidated: both provider accounts ran out of credit mid-run, and it predates the forced-move fix (Appendix~\ref{app:run1}); the whole batch was rerun, and only run~2 is reported here. Neither decision agent made an API error, a malformed or illegal move, or met a \code{tooManyOptions} position (Q9 confirmed; \TimeoutsGem{} and \TimeoutsNim{} timeouts in \NDecGem{} and \NDecNim{} decisions). Both beat \code{random} (\RecGemRandom{} and \RecNimRandom). Beyond that they part ways. \code{gemini-decision} lost every game to \code{rule} (\RecGemRule) and is behind \code{lookahead} (\RecGemLook), as expected. \code{nimble-decision} is not distinguishable from \code{rule} in 50 games (\RecNimRule, \WinNimRule\%; the interval admits both a weaker and an equal player) and beats \code{lookahead} (\RecNimLook), which the expectation ruled out. Against \code{random} both decision agents are better than both typed agents, and against \code{rule} \code{nimble-decision} is too (\WinNimRule\% vs.\ 0\%), so the typed part of Q8 is not confirmed either.

\begin{figure}[t]
\centering
\includegraphics[width=\columnwidth]{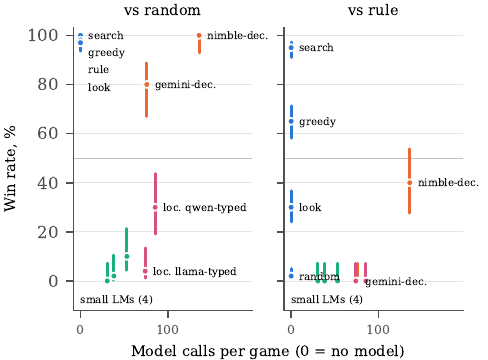}
\caption{Quality against model calls: win rate against \code{random} (left) and \code{rule} (right), 95\% Wilson intervals, against model calls per game (decisions that reached the model; forced moves played without a call are excluded). Reference players and \code{search} make no model call (0). Blue: free players (\code{search}, the ceiling, at the top). Green: small language models. Orange: decision models. Magenta: local agents (batch~6); hollow and dotted: 12-game pairings (indicative).}
\label{fig:cost}
\end{figure}

\paragraph{Quality against model calls.} Figure~\ref{fig:cost} puts the results on the axis that matters for deployment. The players that make no model call span the whole range of quality, from \code{random} to \code{search}, at 0.1--61\,ms per decision. The paid agents make \CallsGameLlamaFree--\CallsGameNim{} model calls per game (Table~\ref{tab:agents}): the language-model agents call the model on every decision (\CallsGameLlamaTyped{} per game for \code{llama8b-typed}), the decision agents on every decision except forced moves (\CallsGameGem{} for \code{gemini-decision}, \CallsGameNim{} for \code{nimble-decision}); each call re-sends the full rules and state, about \TokensGameGem{} input tokens per game for \code{gemini-decision} and \TokensGameNim{} for \code{nimble-decision}, at hundreds of milliseconds per call (median \MedMsGem{} and \MedMsNim\,ms). For those calls the small language models land at or below \code{random}, \code{gemini-decision} between \code{random} and \code{lookahead}, and \code{nimble-decision} not distinguishable from \code{rule} in 50 games. Calling a model on every move buys legality for certain and strength only sometimes; which model is called matters more than the interface.

\paragraph{Search ceiling (batch 5).}\label{sec:search} Amendment~v3 adds \code{search}, a free deterministic rollout search tuned on tuning seeds only, played against the four other references on all 100 test seeds (\NBatchFive{} games, all under protocol v3, 19 Nov 2026). Q10s is \emph{confirmed}: it beats \code{random} \WinSearchRandom\%, \code{rule} \WinSearchRule\%, \code{lookahead} \WinSearchLook\% and \code{greedy-normal} \WinSearchGreedy\% of games, every interval above 50\%, with \BadSearch{} invalid moves at \MsDecSearch\,ms per decision. A free 60\,ms search thus outplays every paid model by the full range of the scale against the same opponents (the two never met directly): against \code{rule} it wins \WinSearchRule\% of games, where the best paid model wins \WinNimRule\% and the others none.

\paragraph{Local agents (batch 6).}\label{sec:local} Protocol v4 serves Llama~3.1 8B and Qwen~2.5 7B locally on one GPU, as free and typed agents with the same prompts and no paid call, against \code{random}, \code{rule} and (typed) \code{search}. The owner cut the run at 6 of 25 seeds (v4.2); v4.3, committed before the extra games, continued only the typed agents against \code{random} and \code{rule}, which now have 50 games each (about $\pm$13 points). The free agents and the games against \code{search} stay at 12 games; they are reported only in Appendix~\ref{app:batch6}. The local typed agents play no invalid move and lose all 100 games to \code{rule}. Against \code{random}, \code{localllama8b-typed} wins \WinLocLlamaTypedRandom\% (\RecLocLlamaTypedRandom{} win--loss--draw), so \code{random} is better: the Q3 refutation repeats locally, and ``better than \code{random}'' in Q13 is refuted for Llama. \code{localqwen7b-typed} wins \WinLocQwenTypedRandom\% (\RecLocQwenTypedRandom), not distinguishable from \code{random}, and better than the local Llama, which refutes Q12's ``not distinguishable'' against \code{random}. The local Llama matches the hosted one on win rates (Q10; \WinLocLlamaTypedRandom\% vs.\ \WinLlamaTypedRandom\% against \code{random}, overlapping). Every game replays from its log (Q11). With one model on the GPU a typed decision takes \SecDecBlockRange\,s (Q16); when games of both models are in flight, Ollama swaps the two models on the 11\,GB GPU and a call takes 5--15\,s, so the run was reordered into one block per model (about 47\,s of wall time per game, against about 55\,s interleaved).

\FloatBarrier
\section{Related Work}

\paragraph{Games as agent benchmarks.} CivRealm~\citep{qi2024civrealm} puts learning and language agents in a full Civilization-like game, where both struggle; hexfolk is deliberately the opposite scale, short enough to play hundreds of times on a held-out seed set. BALROG~\citep{paglieri2025balrog} aggregates existing game environments and reports a gap between what models know and what they do; TextArena~\citep{guertler2025textarena} collects competitive text games rated by TrueSkill; GameBench~\citep{costarelli2024gamebench} chose games unlikely to appear in pretraining data, the concern that led us to write a new game. Our contribution is not breadth but protocol: pre-registered expectations, a held-out seed set used once, mirrored seats, intervals rather than ratings, and a referee that counts invalid moves instead of re-prompting.

\paragraph{Decision models.} Typed decision models return a distribution over caller-defined options in one forward pass~\citep{jev}; open models and data followed~\citep{nimble}, with first-principles analyses~\citep{jeva}, the observation that any causal LLM can be scored the same way~\citep{llmasjev}, and an evaluation checklist~\citep{audit}. \citet{li26agentloops} diagnose their decision failures inside agent loops. MBYT~\citep{lucca2026mbyt} compared them with generative LLMs and a classical baseline on email routing, where options are few and fixed; hexfolk tests them where options are many and change at every step.

\section{Limitations}\label{sec:limits}

\textbf{No frontier model, by budget.} The study works with a limited budget of paid API calls. It covers low-cost decision models and small open language models, not frontier models, so nothing here says how the strongest available models play hexfolk; that comparison is future work, and the harness, seeds and questions are ready for it. \textbf{Statistics.} Paid pairings stay at 25 seeds (50 games) by budget: intervals up to about $\pm$13 points (at a 50\% win rate; $\pm$9--11 near 10--20\%), enough for the large gaps reported here (including the refutation of Q3) but not for small ones, so a ``not distinguishable'' verdict on a paid pairing is weak evidence, not evidence of equality. Invalid-move intervals treat decisions as independent and are optimistic. \textbf{Planning, not inference.} There is no fog of war and combat is deterministic, so the benchmark measures planning under full information. \textbf{One game.} A second game or larger maps may rank agents differently. Its mechanics are inspired by The Battle of Polytopia (economy, techs, small hex map) and Catan (numbered resource tiles); the rules were rewritten and no assets were reused, so general strategic knowledge is available to models and only hexfolk-specific knowledge is absent. \textbf{Few models; an incomplete batch.} Two small language models and two decision models, which already disagree with each other, say little about either class. In batch~6 the free agents and the games against \code{search} rest on 12 games each: only their 0\%/100\% results are firm. \textbf{Option order.} In batches 2--6 the options came in a fixed order with \code{end\_turn} last; the order control (batch~7) covers only the two local typed models against \code{random}. \textbf{Protocol order and prompts.} Batch~1 ran under v1 before v2 was written; the language-model rules text gained one rule on tuning seeds before testing, after batch~4 had run, so the decision agents and the language-model agents did not read identical rules text; no prompt search was done for any model. \textbf{Calls, not money.} We report model calls and input tokens, not money, because provider prices change while call and token counts do not; there is no cost column. \textbf{Infrastructure.} Hosted models are not guaranteed deterministic; for them the log, not the seed, replays a game. The same holds in practice for the local models: on a tuning seed, the same prompt gave a different first choice when another game shared the GPU.

\section{Conclusion}

On hexfolk the MBYT method carries over intact: expectations were written before the games, held-out seeds were used once, the protocol's history is reported rather than smoothed, and every number is generated from the result file. The free players confirmed their expectations and set two bars: a few lines of hand-written priorities beat a one-step search by more than two to one, and a 60\,ms rollout search beats every other reference in at least 89\% of games. The cheap paid agents did not clear it. Two small language models lost every game to it; in free text they were invalid almost every time, and restricted to legal moves they lost even to a random player, refuting a pre-registered expectation, and a control that shuffles the options does not change that. The two decision models never played an invalid move and both beat \code{random}, but they are not interchangeable: \code{gemini-decision} lost 0--50 to \code{rule}, while \code{nimble-decision} won \WinNimRule\% of 50 games against it, not distinguishable in 50 games: these data can say neither that it is weaker than the rule list nor that it matches it. Neither decision model is shown better than \code{rule}; both take a model call of hundreds of milliseconds per move against well under a millisecond, and, against the same opponents, the free search outplays them all by the full range. Two 7--8B models served locally point the same way: typed, the Llama model loses to \code{random} again, and the Qwen model is not distinguishable from it. This is a narrow result on one game and says nothing about frontier models. What it shows is that, for a model call per move, legality is not strength, the simple free baseline is the bar to measure first, and the choice of model matters as much as the choice of interface.

\section*{Code and Data Availability}
Code and protocol: \url{https://github.com/federicolucca/hexfolk}, tag \code{bench-v1} (the engine, the benchmark harness and its agents, the tests, and the pre-registered protocol \code{docs/PROTOCOL.md} with its version history v1--v5; benchmark command \code{npm run bench}). The per-game results and the game logs are not in the repository; they are available from the author on request. Every game replays from its log with the released code. Hashes refer to the public repository; the per-game results were removed from its history before release and are available on request. Appendix~\ref{app:history} maps each protocol version to its commit.

\section*{Use of AI Assistants}
The research idea, the design of the game, the pre-registered questions and the interpretation of the results are the author's. AI coding assistants gave practical support: implementing the code and scripts, running the benchmark batches, writing the analysis scripts, and helping draft and edit the text under the author's direction. The author checked the results and takes full responsibility for the content of this paper.

\bibliography{refs}

\begin{thebibliography}{16}
\providecommand{\natexlab}[1]{#1}

\bibitem[{Almeida(2026)}]{jev}
D.~Almeida. 2026.
\newblock Introducing system one models \& {Jev}.
\newblock TypeSafe AI blog, 15 September 2026.

\bibitem[{{Anthropic}(2024)}]{anthropic2024mcp}
{Anthropic}. 2024.
\newblock Introducing the {Model Context Protocol}.
\newblock \url{https://www.anthropic.com/news/model-context-protocol}.

\bibitem[{{Bespoke Labs} et~al.(2026){Bespoke Labs}, Pimpalgaonkar, Dalal, and
  Sathiamoorthy}]{nimble}
{Bespoke Labs}, S.~Pimpalgaonkar, P.~Dalal, and M.~Sathiamoorthy. 2026.
\newblock Nimble: data, model, recipe for an open {Jev}.
\newblock \url{https://github.com/bespokelabsai/nimble}.

\bibitem[{Choi and Yu(2026)}]{choi2026gemini}
Min Choi and Felix Yu. 2026.
\newblock Gemini decisions {API}: Fast, calibrated multimodal decisions.
\newblock Google blog, 11 November 2026.
  \url{https://blog.google/innovation-and-ai/technology/developers-tools/gemini-decisions-api/}.

\bibitem[{Costarelli et~al.(2024)}]{costarelli2024gamebench}
Anthony Costarelli et~al. 2024.
\newblock {GameBench}: Evaluating strategic reasoning abilities of {LLM}
  agents.
\newblock arXiv:2406.06613. NeurIPS 2024 Workshop on Language Gamification.

\bibitem[{Guertler et~al.(2025)Guertler, Cheng, Yu, Liu, Choshen, and
  Tan}]{guertler2025textarena}
Leon Guertler, Bobby Cheng, Simon Yu, Bo~Liu, Leshem Choshen, and Cheston Tan.
  2025.
\newblock {TextArena}.
\newblock arXiv:2504.11442.

\bibitem[{Li and Wang(2026)}]{jeva}
Q.~Li and X.~Wang. 2026.
\newblock {Jeva}: System one models from first principles.
\newblock arXiv:2610.18773.

\bibitem[{Li et~al.(2026)Li, Krishnan, and Padman}]{li26agentloops}
Y.~Li, R.~Krishnan, and R.~Padman. 2026.
\newblock Reading right, acting wrong: diagnosing decision failures of system
  one models in agent loops.
\newblock arXiv:2610.18579.

\bibitem[{Li and Wagle(2026)}]{llmasjev}
Y.~Li and J.~Wagle. 2026.
\newblock {LLM}-as-{Jev}: {LLMs} are already {Jev}-style decision models---when
  and how to fine-tune them.
\newblock arXiv:2610.02076.

\bibitem[{Lucca(2026)}]{lucca2026mbyt}
Federico Lucca. 2026.
\newblock Measure before you train: Typed decision models vs.\ a classical
  baseline vs.\ a generative {LLM} for routing a personal multilingual inbox.
\newblock arXiv:2610.32955.
\newblock Companion paper.

\bibitem[{Nosek et~al.(2018)Nosek, Ebersole, DeHaven, and
  Mellor}]{nosek2018prereg}
Brian~A. Nosek, Charles~R. Ebersole, Alexander~C. DeHaven, and David~T. Mellor.
  2018.
\newblock The preregistration revolution.
\newblock \emph{Proceedings of the National Academy of Sciences},
  115(11):2600--2606.

\bibitem[{Paglieri et~al.(2025)}]{paglieri2025balrog}
Davide Paglieri et~al. 2025.
\newblock {BALROG}: Benchmarking agentic {LLM} and {VLM} reasoning on games.
\newblock In \emph{The Thirteenth International Conference on Learning
  Representations (ICLR)}.
\newblock ArXiv:2411.13543.

\bibitem[{Qi et~al.(2024)}]{qi2024civrealm}
Siyuan Qi et~al. 2024.
\newblock {CivRealm}: A learning and reasoning odyssey in {Civilization} for
  decision-making agents.
\newblock In \emph{The Twelfth International Conference on Learning
  Representations (ICLR)}.
\newblock ArXiv:2401.10568.

\bibitem[{Sainz et~al.(2023)Sainz, Campos, Garc{\'i}a-Ferrero, Etxaniz,
  Lopez~de Lacalle, and Agirre}]{sainz2023contamination}
Oscar Sainz, Jon~Ander Campos, Iker Garc{\'i}a-Ferrero, Julen Etxaniz, Oier
  Lopez~de Lacalle, and Eneko Agirre. 2023.
\newblock {NLP} evaluation in trouble: On the need to measure {LLM} data
  contamination for each benchmark.
\newblock In \emph{Findings of the Association for Computational Linguistics:
  EMNLP 2023}.

\bibitem[{Tang and Zheng(2026)}]{audit}
L.~Tang and Y.~Zheng. 2026.
\newblock Typed decision models: An early evidence audit and evaluation
  checklist.
\newblock arXiv:2609.32160.

\bibitem[{Wilson(1927)}]{wilson1927}
Edwin~B. Wilson. 1927.
\newblock Probable inference, the law of succession, and statistical inference.
\newblock \emph{Journal of the American Statistical Association},
  22(158):209--212.

\end{thebibliography}

\appendix

\section{Protocol history}\label{app:history}

\begin{table*}[htbp]
\centering\small
\caption{Protocol versions and their commits in the public repository (\url{https://github.com/federicolucca/hexfolk}, \code{main}). Dates and times are the commit times (UTC+1), verified on \code{main} (tip \code{9968404}, tag \code{bench-v1}); the history after the run-1 fix was rewritten before release to remove the per-game results, keeping the dates.}
\label{tab:versions}
\begin{tabular}{llll}
\toprule
Version & Committed & Commit & Related code commits \\
\midrule
v1 & 19 Nov 2026, 08:43 & \code{e2379b7} & harness \code{ec33e37} \\
v2 & 19 Nov 2026, 09:10 & \code{8753b19} & -- \\
v2.1 & 19 Nov 2026, 10:56 & \code{6e89ee8} & \code{f44cad3}, \code{483e79c}, \code{b523556}, \code{ece32ad} \\
v3 & 19 Nov 2026, 12:23 & \code{953787a} & \code{search}: \code{2b321bf} \\
v4 & 20 Nov 2026, 10:45 & \code{15ef8b3} & local agents: \code{28723aa} \\
v4.1 & 20 Nov 2026, 11:16 & \code{5a24573} & -- \\
v4.2 & 20 Nov 2026, 13:15 & \code{832a4eb} & -- \\
v4.3 & 20 Nov 2026, 13:21 & \code{d39d735} & -- \\
v5 & 20 Nov 2026, 16:17 & \code{2087862} & shuffled agent: \code{801009b} \\
\bottomrule
\end{tabular}
\end{table*}

\paragraph{Version 1} (commit \code{e2379b7}, 19 November 2026) was committed before any final result. It fixes the agents and models, the harness rules, the seeds and three final batches (1--3). Tuning seeds 1--50 had already been used for a reference table (600 games) and seeds 1--2 for language-model pilots. Statistics as in Section~\ref{sec:protocol}; timeouts and errors are reported apart, and an agent with more than 2\% errors is flagged.

\paragraph{Amendment v2} (commit \code{8753b19}) was written the same day, \emph{after batch~1 had been run under v1} and before any decision-model game on the test set; batches 2 and 3 had not been run and were unchanged. It adds the \code{decision} agent type, batch~4 and Q8--Q9. Tuning seeds could be used only to fix the wording of the decision agent's instruction (pilots: 4 games per model on seeds 1--2 against \code{rule}).

\paragraph{Version 2.1} (commit \code{6e89ee8}, documentation only) records, before any language-model game on the test set, four changes made after v2. (i) A forced move (one legal action) is played without a model call, because one provider refuses a one-option question; no move changes (commit \code{f44cad3}). (ii) A batch stops at the first HTTP~402 (out of credit) and does not record that game (commit \code{483e79c}); batch~4 run~1 was invalid for that reason (Appendix~\ref{app:run1}). (iii) A run can be given a hard spending limit; a batch cut by it is reported as incomplete. (iv) A \emph{pre-test prompt fix found on tuning seeds} (commit \code{ece32ad}): the shared rules text given to the language-model agents now states that a city can train a unit only when its tile is empty. In the pilot on tuning seeds 1--2 (16 games), 150 of the free agents' 239 illegal moves were attempts to train in an occupied city, a rule the text had never stated; a re-pilot on tuning seed~1 still showed only invalid moves for the free agents, and nothing further was changed. Batch~2 ran on \code{ece32ad} and batch~3 on \code{6e89ee8} (same code). Both batch-4 runs predate the fix (run~2 ran on \code{483e79c}).

\paragraph{Budget.} The study works with a limited budget of paid API calls. Batches 2 and 3 ran under a hard stop on paid calls. Paid pairings therefore stay at 25 seeds and no frontier model is run; free pairings may be extended to more seeds.

\paragraph{Amendment v3} (commit \code{953787a}) was written after batches 1--4 had run and before any test game of the new player. It adds \code{search}: a deterministic rollout search that plays each shortlisted action 5 turns forward, with the game's AI on its side and the worse of the game's AI and \code{rule} on the other, under a fixed node budget, tuned on tuning seeds only and frozen. Batch~5 plays it against the four other references on all 100 test seeds, both seats (800 games, no paid calls). Q10s: \code{search} is better than each reference, with its win-rate interval above 50\% in every pairing.

\paragraph{Amendment v4} (commit \code{15ef8b3}) was written before any local game. It adds local agents served by Ollama on one RTX 2080 Ti (\code{localllama8b} = \code{llama3.1:8b}, \code{localqwen7b} = \code{qwen2.5:7b}, both 4-bit), each as a free and a typed agent with the prompts of \code{ece32ad}, and questions Q10--Q16; Q15 (a local decision model) was dropped before testing because no local server for it exists. \textbf{Version 4.1} (commit \code{5a24573}) fixed, after a pilot on tuning seeds 1--2 and before any test game, the 60\,s timeout and batch~6: the four agents against \code{random} and \code{rule} and the two typed agents against \code{search}, on seeds 100001--100025, both seats (500 games); the direct match against hosted \code{llama8b} and the hosted replays of Q11 were skipped because they would need paid calls. \textbf{Version 4.2} (commit \code{832a4eb}, written after the run) records that the owner cut batch~6 after 126 games: seeds 100001--100006 are kept (120 games, 12 per pairing), the partial seed 100007 is dropped, Q11 is reduced to log replays, and the batch is reported as incomplete. \textbf{Version 4.3} (commit \code{d39d735}, written before the extra games) continued batch~6 on seeds 100007--100025 for the typed agents only, against \code{random} and \code{rule}, both seats (152 games; seed 100007 replayed in full), at the owner's request for firmer numbers at the lowest GPU cost. The free agents and the games against \code{search} were not continued, because their 12 games were already one-sided (0 wins in each pairing). The continuation ran on code identical to that of v4.1 (\code{5a24573}): its logs carry later commit identifiers from before release, which changed only documentation. \textbf{Amendment v5} (commit \code{2087862}, 16:17 on 20 November; the first test game of batch~7 started at 16:24) adds an option-order control, requested after this paper's review found that \code{end\_turn} is always the last option. A new agent, \code{<model>-typed-shuffled} (code \code{801009b}), uses the typed prompt, schema and settings, but at each decision shows the legal list in a permutation drawn from the game seed and the decision index; the permutation is logged and the choice mapped back. Batch~7 plays the two local typed agents, shuffled, against \code{random} on seeds 100001--100025, both seats (100 games, no paid call; pilot on tuning seeds 1--2). Q17 expected the \code{end\_turn} share to fall by at least 10 points for both models; Q18 expected the win rate against \code{random} to be not distinguishable from batch~6; Q19 (exploratory) compares the shares of the first and last index with the uniform rate in both orders. The expectations were written by the author of this paper and committed before any shuffled test game. The games ran on the v5 commit, which adds to \code{801009b} only the \code{bench positions} analysis command.

\section{Batch 4, run 1 (invalid)}\label{app:run1}

The first run of batch~4 was declared invalid. The OpenRouter account ran out of credit during the run, so a share of \code{gemini-decision}'s decisions failed and were played, as the referee prescribes, as \code{end\_turn}; the Bespoke account behind \code{nimble-decision} ran out of credit in the same way, and that run was stopped by hand. The run also predates the forced-move fix (\code{f44cad3}): a position whose only legal action is \code{end\_turn} was sent as a one-option question, which both providers reject. It is recorded in the benchmark's run records (available on request with the other per-game data); its raw files are kept apart from the final results, and its numbers are not results. The whole batch was rerun (run~2, Section~\ref{sec:results}); the protocol requires both runs to be reported, run~1 by this description. Since run~1 the benchmark stops a batch at the first HTTP~402 and does not record that game.

\section{Per-agent statistics and win rates}\label{app:more}

\begin{table*}[htbp]
\centering\footnotesize
\setlength{\tabcolsep}{3pt}
\caption{Per agent, reference players and hosted models: decisions, bad decisions (malformed + illegal, \% of decisions; optimistic interval), API errors, timeouts, mean ms per decision (forced moves included), model calls per game (decisions that reached the model; forced moves played without a call are excluded; 0 for players that use no model) and input tokens per game. \code{search}: batch~5; other references: batch~1; language models: batches 2--3; decision models: batch~4 run~2. Local models: Table~\ref{tab:agents-local}.}
\label{tab:agents}
\begin{tabular}{lrrcrrrrr}
\toprule
Agent & Games & Decisions & Bad \% [95\% CI] & Err. & T/O & ms/dec. & Calls/game & Tok.\ in/game \\
\midrule
\code{search} & 800 & 96{,}448 & 0.00 [0.00, 0.00] & 0 & 0 & 60.7 & 0 & -- \\
\code{greedy-normal} & 600 & 63{,}547 & 0.00 [0.00, 0.01] & 0 & 0 & 0.2 & 0 & -- \\
\code{rule} & 600 & 55{,}740 & 0.00 [0.00, 0.01] & 0 & 0 & 0.1 & 0 & -- \\
\code{lookahead} & 600 & 57{,}373 & 0.00 [0.00, 0.01] & 0 & 0 & 2.6 & 0 & -- \\
\code{random} & 600 & 43{,}908 & 0.00 [0.00, 0.01] & 0 & 0 & 0.1 & 0 & -- \\
\code{llama8b-typed} & 150 & 7{,}964 & 0.00 [0.00, 0.05] & 0 & 1 & 544.8 & 53.1 & 70k \\
\code{ministral3b-typed} & 150 & 5{,}737 & 0.00 [0.00, 0.07] & 0 & 0 & 476.2 & 38.2 & 54k \\
\code{llama8b-free} & 150 & 4{,}565 & 97.8 [97.4, 98.2] & 0 & 0 & 1{,}433.8 & 30.4 & 38k \\
\code{ministral3b-free} & 150 & 4{,}585 & 97.8 [97.3, 98.2] & 0 & 0 & 507.0 & 30.6 & 41k \\
\code{gemini-decision} & 150 & 11{,}459 & 0.00 [0.00, 0.03] & 0 & 4 & 506.2 & 75.8 & 147k \\
\code{nimble-decision} & 150 & 21{,}706 & 0.00 [0.00, 0.02] & 0 & 2 & 408.4 & 135.6 & 299k \\
\bottomrule
\end{tabular}

\end{table*}

\begin{table*}[htbp]
\centering\footnotesize
\setlength{\tabcolsep}{3pt}
\caption{Per agent, local models (columns as in Table~\ref{tab:agents}): batch~6 (typed agents: 112 games; free agents: 24, indicative) and batch~7 ($^{s}$: shuffled option order, 50 games each).}
\label{tab:agents-local}
\begin{tabular}{lrrcrrrrr}
\toprule
Agent & Games & Decisions & Bad \% [95\% CI] & Err. & T/O & ms/dec. & Calls/game & Tok.\ in/game \\
\midrule
\code{localllama8b-typed} & 112 & 8{,}285 & 0.00 [0.00, 0.05] & 0 & 0 & 1{,}589.8 & 74.0 & 95k \\
\code{localqwen7b-typed} & 112 & 9{,}565 & 0.00 [0.00, 0.04] & 0 & 0 & 1{,}432.3 & 85.4 & 113k \\
\code{localllama8b-free} & 24 & 720 & 100.0 [99.5, 100.0] & 0 & 0 & 1{,}414.6 & 30.0 & 38k \\
\code{localqwen7b-free} & 24 & 731 & 98.5 [97.3, 99.2] & 0 & 0 & 2{,}298.4 & 30.5 & 39k \\
\code{localllama8b-typed}$^{s}$ & 50 & 3{,}794 & 0.00 [0.00, 0.10] & 0 & 0 & 1{,}259.3 & 75.9 & 102k \\
\code{localqwen7b-typed}$^{s}$ & 50 & 3{,}213 & 0.00 [0.00, 0.12] & 0 & 0 & 1{,}149.9 & 64.3 & 89k \\
\bottomrule
\end{tabular}

\end{table*}

\begin{table*}[htbp]
\centering\footnotesize
\caption{Decision agents in the logs of batch~4 run~2. Calls exclude forced moves. Mean $p$ per game: the average probability the model gave to its chosen action, averaged within a game, by the game's outcome for the decision agent.}
\label{tab:decision}
\begin{tabular}{lrr}
\toprule
 & \code{gemini-decision} & \code{nimble-decision} \\
\midrule
Model calls answered & 11{,}367 & 20{,}337 \\
Forced moves (no call) & 88 & 1{,}367 \\
API errors & 0 & 0 \\
Timeouts ($>$60\,s) & 4 & 2 \\
\code{tooManyOptions} & 0 & 0 \\
Median ms per call & 367 & 348 \\
90th pct.\ ms per call & 577 & 443 \\
Mean $p$ of chosen action & 0.218 & 0.156 \\
Mean $p$ per game, wins (games) & 0.183 (51) & 0.150 (102) \\
Mean $p$ per game, losses (games) & 0.243 (97) & 0.205 (45) \\
Mean $p$ per game, draws (games) & 0.209 (2) & 0.159 (3) \\
\bottomrule
\end{tabular}

\end{table*}

\paragraph{Decision agents in detail.} \code{gemini-decision} made \NDecGem{} decisions over its 150 games and \code{nimble-decision} \NDecNim, so their model calls (\CallsTotalGem{} and \CallsTotalNim; \ForcedTotalGem{} and \ForcedTotalNim{} forced moves were played without a call) and input tokens, which scale with the number of actions, differ by about a factor of two. \emph{Confidence (exploratory, no expectation).} The probability a model gives its chosen action is low---\MeanPGem{} and \MeanPNim{} on average---and is \emph{not} higher in games the agent wins (\PWinGem{} vs.\ \PLossGem{} for \code{gemini-decision}, \PWinNim{} vs.\ \PLossNim{} for \code{nimble-decision}). The opponent is a plausible confound; a per-position analysis is needed before this says anything about calibration.

\begin{figure*}[htbp]
\centering
\includegraphics[width=\textwidth]{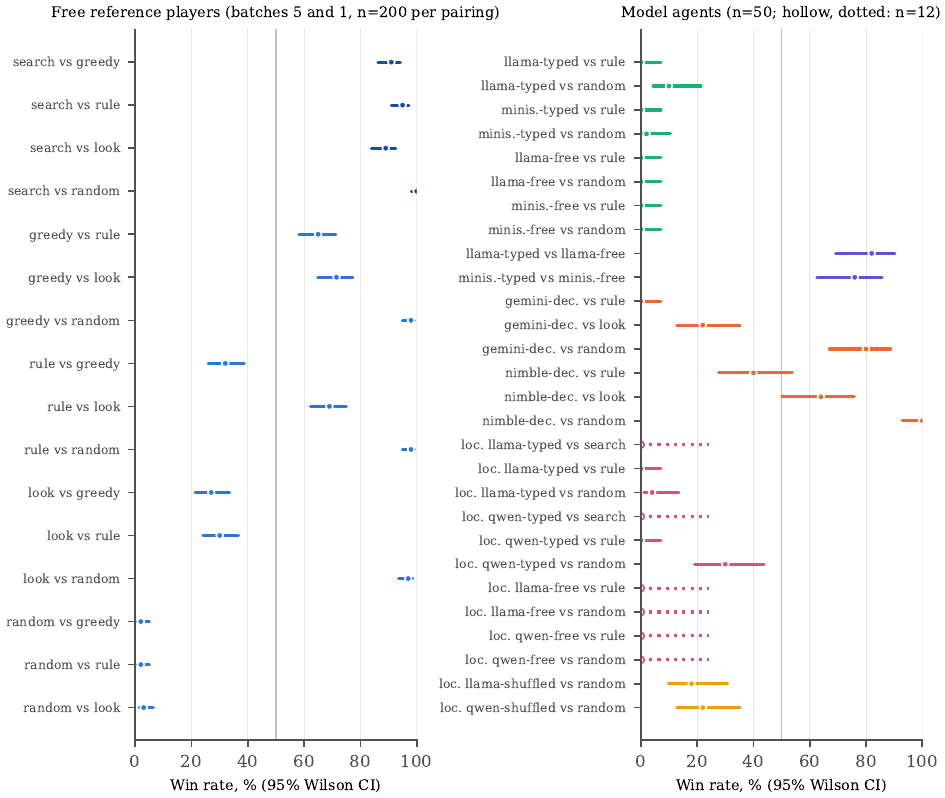}
\caption{Win rates with 95\% Wilson intervals, both seats pooled. Left: \code{search} against each reference (batch~5), then each other reference player against each other one (batch~1). Right: each model agent against its opponents (batch~2 green, batch~3 violet, batch~4 orange, batch~6 magenta; hollow with dotted intervals: 12-game pairings, indicative).}
\label{fig:win}
\end{figure*}

\section{Batch 6: local agents (incomplete)}\label{app:batch6}

\begin{table*}[htbp]
\centering\footnotesize
\caption{Batch~6: local agents (Ollama, one RTX 2080 Ti, no paid call), both seats. Typed agents against \code{random} and \code{rule}: seeds 100001--100025, 50 games. $^\dagger$12 games (seeds 100001--100006, about $\pm$25 points): indicative only. Code: protocol v4.1, commit \code{5a24573}.}
\label{tab:batch6}
\begin{tabular}{llrcccl}
\toprule
A & B & $n$ & A wins \% [95\% CI] & B wins \% [95\% CI] & D & Verdict \\
\midrule
\code{localllama8b-typed} & \code{random} & 50 & 4.0 [1.1, 13.5] & 60.0 [46.2, 72.4] & 18 & \code{random} better \\
\code{localllama8b-typed} & \code{rule} & 50 & 0.0 [0.0, 7.1] & 100.0 [92.9, 100.0] & 0 & \code{rule} better \\
\code{localllama8b-typed} & \code{search} & 12$^\dagger$ & 0.0 [0.0, 24.2] & 100.0 [75.8, 100.0] & 0 & \code{search} better \\
\code{localqwen7b-typed} & \code{random} & 50 & 30.0 [19.1, 43.8] & 40.0 [27.6, 53.8] & 15 & n.d. \\
\code{localqwen7b-typed} & \code{rule} & 50 & 0.0 [0.0, 7.1] & 100.0 [92.9, 100.0] & 0 & \code{rule} better \\
\code{localqwen7b-typed} & \code{search} & 12$^\dagger$ & 0.0 [0.0, 24.2] & 100.0 [75.8, 100.0] & 0 & \code{search} better \\
\code{localllama8b-free} & \code{random} & 12$^\dagger$ & 0.0 [0.0, 24.2] & 100.0 [75.8, 100.0] & 0 & \code{random} better \\
\code{localllama8b-free} & \code{rule} & 12$^\dagger$ & 0.0 [0.0, 24.2] & 100.0 [75.8, 100.0] & 0 & \code{rule} better \\
\code{localqwen7b-free} & \code{random} & 12$^\dagger$ & 0.0 [0.0, 24.2] & 100.0 [75.8, 100.0] & 0 & \code{random} better \\
\code{localqwen7b-free} & \code{rule} & 12$^\dagger$ & 0.0 [0.0, 24.2] & 100.0 [75.8, 100.0] & 0 & \code{rule} better \\
\bottomrule
\end{tabular}

\end{table*}

Batch~6 ran in two parts on 20 November 2026 (\NBatchSix{} games, no timeout, no error): seeds 100001--100006 for every pairing (11:17--13:15), where the owner cut the planned 500-game batch (v4.2; the 6 games of the partial seed 100007 are dropped), and seeds 100007--100025 for the typed agents against \code{random} and \code{rule} (13:35--15:51, v4.3). Per-agent figures (invalid moves, latency, model calls and input tokens per game) are in Table~\ref{tab:agents}. The free agents are invalid on \BadLocLlamaFree{} of \NDecLocLlamaFree{} (\code{localllama8b-free}) and \BadLocQwenFree{} of \NDecLocQwenFree{} (\code{localqwen7b-free}) decisions; the typed agents on none of \NDecLocLlamaTyped{} and \NDecLocQwenTyped. All \ReplaySixGames{} games rebuild from their logs with the logged winner and scores (\ReplaySixMismatches{} mismatches) through the engine copy of Appendix~\ref{app:actions}; hexfolk itself replayed 2. Replay from the seed was not tested on the test set, and a tuning-seed pair in the pilot (same prompt, a different first choice when another game shared the GPU) contradicts it. \emph{Latency.} Mean time per decision over the batch is \SecDecLocLlamaTyped{} and \SecDecLocQwenTyped\,s for the typed agents and \SecDecLocLlamaFree{} and \SecDecLocQwenFree\,s for the free ones, against about 0.5\,s for the hosted typed agents. It depends on scheduling: the GPU (11\,GB) holds one model at a time, so with games of both models in flight Ollama swaps them and a call takes 5--15\,s, against about 1\,s otherwise. The first part interleaved the two models (126 games in about 116 minutes, about 55\,s of wall time per game); from 14:00 each model's games ran as one block (141 games in 111 minutes, about 47\,s per game), and on seeds 100007--100025 a typed decision took \SecDecBlockLocLlamaTyped{} and \SecDecBlockLocQwenTyped\,s. The run times are from hexfolk's run records; the per-decision times are from the result file.

\section{Action mix, actions per turn and option position (exploratory)}\label{app:actions}

The typed agents end their turn early: \code{end\_turn} is \ActLlamaTypedEndturn\% of \code{llama8b-typed}'s actions and \ActMinTypedEndturn\% of \code{ministral3b-typed}'s, against \ActRandomVsTypedEndturn\% for \code{random} in the same games, so they take \DecGameLlamaTyped{} and \DecGameMinTyped{} actions per game against \code{random}'s \DecGameRandomVsTyped. They move little (\ActLlamaTypedMove\% and \ActMinTypedMove\% of actions, against \ActRandomVsTypedMove\%) and hold more stars when they pass: a median of \MedUnspentLlamaTyped{} and \MedUnspentMinTyped{} unspent stars at the end of a turn, against \MedUnspentRandomVsTyped{} for \code{random}, which spends whatever it has. In a game won by expanding and capturing, an agent that passes with its stars in hand and its units at home falls behind even a random opponent that keeps acting. The decision agents look different, \code{nimble-decision} most of all (\ActNimEndturn\% \code{end\_turn}, \DecGameNim{} actions per game, a median of \MedUnspentNim{} unspent stars). The free agents' rows restate their invalid rate (\ActLlamaFreeEndturn\% of applied actions are \code{end\_turn}). The replay uses a read-only copy of the engine at commit \code{953787a} (rules v1.0.0). This reading is a hypothesis for a future protocol---for example, an expectation on the end-turn share---not a finding of this one.

\paragraph{Option position and actions per turn.} In batches 2--6 the harness numbers the legal actions in the engine's order (moves of each unit first, then captures, attacks, research, walls, roads, training, harvests) and appends \code{end\_turn} last; nothing shuffles the list (checked at the code commits of batches 2, 4 and 6), so \code{end\_turn} is the last option in all \PosDecisions{} decisions we rebuilt. Table~\ref{tab:positions} gives the position shares over all valid decisions (a one-option list counts as both first and last), as hexfolk's \code{bench positions} computes them; Table~\ref{tab:positions-bins} splits them by list length. Two effects are mixed in the engine order. One is a real primacy bias: Llama takes the first option in \PosFirstDLocLlamaTyped\% of its local decisions against \PosUnifLocLlamaTyped\% for a uniform choice, and keeps it when the order is shuffled (\PosFirstDLocLlamaShuf\% against \PosUnifLocLlamaShuf\%); in engine order the first option is usually a move of the first-listed unit. The other only looks like a bias for the last option: because \code{end\_turn} is last, its share appears as ``last option'' in engine order, but that share is set by the number of actions per turn (Table~\ref{tab:apt}), and when the list is shuffled the last shown option falls below uniform (\PosLastDLocLlamaShuf\% and \PosLastDLocQwenShuf\%) while the \code{end\_turn} share does not fall. The batch-7 control (Section~\ref{sec:order}) shows that neither effect is why the typed agents lose to \code{random}. \code{random} matches the uniform shares, which checks the reconstruction.

\begin{table*}[htbp]
\centering\footnotesize
\caption{Exploratory (not pre-registered): what the paid agents do with their decisions, from \ReplayGames{} batch-2/3/4 logs replayed through the engine (\ReplayMismatches{} replays disagree with the logged result). Decisions per game; share of applied actions (an invalid decision is applied as \code{end\_turn}; ``other'' is research, harvest, capture, roads and walls); stars held when the agent ends its turn, mean and median. The last two rows are the opponents' seats in the typed agents' batch-2 games.}
\label{tab:actions}
\begin{tabular}{lrrrrrrrr}
\toprule
Agent & Dec./game & end\_turn & move & train & attack & other & \multicolumn{2}{c}{Unspent $\star$ mean / median} \\
\midrule
\code{llama8b-typed} & 53 & 52\% & 28\% & 3\% & 1\% & 15\% & 25.7 & 20 \\
\code{ministral3b-typed} & 38 & 71\% & 11\% & 3\% & 3\% & 12\% & 36.7 & 32 \\
\code{llama8b-free} & 30 & 98\% & 0\% & 0\% & 0\% & 2\% & 33.1 & 33 \\
\code{ministral3b-free} & 31 & 98\% & 1\% & 1\% & 0\% & 1\% & 34.0 & 33 \\
\code{gemini-decision} & 76 & 36\% & 33\% & 8\% & 9\% & 14\% & 35.9 & 20 \\
\code{nimble-decision} & 145 & 17\% & 39\% & 11\% & 14\% & 20\% & 22.3 & 1 \\
\code{random} (vs typed) & 114 & 26\% & 57\% & 3\% & 1\% & 13\% & 20.0 & 6 \\
\code{rule} (vs typed) & 100 & 21\% & 53\% & 6\% & 2\% & 18\% & 56.2 & 18 \\
\bottomrule
\end{tabular}

\end{table*}

\begin{table*}[htbp]
\centering\footnotesize
\setlength{\tabcolsep}{3pt}
\caption{Exploratory: position of the chosen option, over all valid decisions that reached the model (forced decision-agent moves excluded), rebuilt by replaying the logs (every reply index maps to the applied action). In engine order \code{end\_turn} is always the last option; rows marked $^{s}$ (batch~7, shuffled order) count positions as shown. Uniform: mean of $1/n$. Non-end: among choices other than \code{end\_turn}, the share in the first and last tenth of the remaining list. Typed: hosted batches 2--3, local batch~6 (all opponents, and vs.\ \code{random} only), batch~7 shuffled; decision agents: batch~4 run~2; \code{random}: the control.}
\label{tab:positions}
\begin{tabular}{lrrrrrrr}
\toprule
Agent & Dec. & Median $n$ & \code{end\_turn} & First shown & Last shown & Uniform & Non-end: first / last 10\% \\
\midrule
\code{llama8b-typed} & 7{,}963 & 12 & 51.7\% & 53.2\% & 51.7\% & 21.2\% & 87\% / 7\% \\
\code{ministral3b-typed} & 5{,}737 & 15 & 70.8\% & 3.1\% & 70.8\% & 9.9\% & 15\% / 33\% \\
\code{localllama8b-typed} & 8{,}285 & 8 & 32.5\% & 72.5\% & 32.5\% & 29.5\% & 80\% / 3\% \\
\code{localqwen7b-typed} & 9{,}565 & 9 & 28.7\% & 40.8\% & 28.7\% & 26.0\% & 35\% / 6\% \\
\quad vs.\ \code{random} only & 3{,}919 & 10 & 38.3\% & 66.2\% & 38.3\% & 27.1\% & 80\% / 4\% \\
\quad vs.\ \code{random} only & 5{,}470 & 9 & 27.3\% & 42.5\% & 27.3\% & 24.6\% & 38\% / 5\% \\
\code{localllama8b-typed}$^{s}$ & 3{,}794 & 12 & 39.5\% & 62.4\% & 15.5\% & 22.1\% & 80\% / 3\% \\
\code{localqwen7b-typed}$^{s}$ & 3{,}213 & 13 & 46.7\% & 21.4\% & 10.4\% & 16.7\% & 23\% / 4\% \\
\code{gemini-decision} & 11{,}367 & 22 & 35.0\% & 8.5\% & 35.0\% & 6.2\% & 18\% / 13\% \\
\code{nimble-decision} & 20{,}337 & 33 & 11.1\% & 10.5\% & 11.1\% & 5.7\% & 21\% / 15\% \\
\code{random} (vs.\ typed agents) & 22{,}859 & 10 & 26.2\% & 26.6\% & 26.2\% & 26.5\% & 16\% / 16\% \\
\bottomrule
\end{tabular}

\end{table*}

\begin{table*}[htbp]
\centering\footnotesize
\caption{Exploratory: share of decisions (\%) choosing the first option (index 0) and the last option (index $n{-}1$, always \code{end\_turn}) by list length $n$, against a uniform choice (mean of $1/n$). Cells with fewer than 30 decisions are omitted.}
\label{tab:positions-bins}
\begin{tabular}{lrrrrrrrrr}
\toprule
 & \multicolumn{3}{c}{$n$ = 2--10} & \multicolumn{3}{c}{$n$ = 11--30} & \multicolumn{3}{c}{$n > 30$} \\
Agent & First & Last & Unif. & First & Last & Unif. & First & Last & Unif. \\
\midrule
\code{llama8b-typed} & 85 & 9 & 20 & 27 & 66 & 7 & -- & -- & -- \\
\code{ministral3b-typed} & 6 & 47 & 16 & 0 & 74 & 7 & -- & -- & -- \\
\code{localllama8b-typed} & 81 & 10 & 20 & 49 & 27 & 7 & -- & -- & -- \\
\code{localqwen7b-typed} & 42 & 27 & 19 & 16 & 3 & 7 & 12 & 5 & 3 \\
\quad vs.\ \code{random} only & 81 & 11 & 19 & 42 & 38 & 7 & -- & -- & -- \\
\quad vs.\ \code{random} only & 47 & 27 & 18 & 19 & 4 & 7 & 12 & 6 & 3 \\
\code{localllama8b-typed}$^{s}$ & 72 & 7 & 20 & 49 & 2 & 6 & 43 & 0 & 3 \\
\code{localqwen7b-typed}$^{s}$ & 29 & 9 & 18 & 9 & 0 & 7 & 4 & 0 & 3 \\
\code{gemini-decision} & 12 & 62 & 15 & 11 & 28 & 6 & 3 & 31 & 2 \\
\code{nimble-decision} & 23 & 25 & 19 & 13 & 12 & 6 & 5 & 7 & 2 \\
\code{random} (vs.\ typed agents) & 20 & 20 & 20 & 7 & 6 & 6 & 3 & 2 & 3 \\
\bottomrule
\end{tabular}

\end{table*}

\begin{table*}[htbp]
\centering\small
\caption{Actions per turn (applied actions other than \code{end\_turn}, per \code{end\_turn}) and the resulting \code{end\_turn} share, from the replayed logs. Each turn ends with exactly one \code{end\_turn} (1{,}500 per 50 games when a game runs its 30 rounds).}
\label{tab:apt}
\begin{tabular}{lrrr}
\toprule
Agent & Games & Actions per turn & \code{end\_turn} share \\
\midrule
\code{llama8b-typed} (hosted) & 150 & 0.93 & 51.7\% \\
\code{ministral3b-typed} (hosted) & 150 & 0.41 & 70.8\% \\
\code{random} vs.\ hosted typed & 100 & 2.80 & 26.3\% \\
\code{localllama8b-typed} vs.\ \code{random} & 50 & 1.61 & 38.3\% \\
\code{localqwen7b-typed} vs.\ \code{random} & 50 & 2.67 & 27.3\% \\
\code{random} vs.\ local typed & 100 & 2.83 & 26.1\% \\
\code{localllama8b-typed}, shuffled & 50 & 1.53 & 39.5\% \\
\code{localqwen7b-typed}, shuffled & 50 & 1.14 & 46.7\% \\
\code{random} vs.\ shuffled & 100 & 2.82 & 26.2\% \\
\code{gemini-decision} & 150 & 1.81 & 35.5\% \\
\code{nimble-decision} & 150 & 4.98 & 16.7\% \\
\bottomrule
\end{tabular}

\end{table*}

\end{document}